\documentclass[11pt]{article}

\usepackage[preprint]{acl}

\usepackage{times}
\usepackage{latexsym}

\usepackage{natbib}

\usepackage[T1]{fontenc}

\usepackage[utf8]{inputenc}

\usepackage{microtype}

\usepackage{inconsolata}

\usepackage{graphicx}
\usepackage{amssymb}
\usepackage{amsmath}
\usepackage{booktabs}
\usepackage{multirow}
\usepackage{bbding}
\title{Benchmarking Text-to-Python against Text-to-SQL: The Impact of Explicit Logic and Ambiguity}

\author{Hangle Hu, Chenyu Hou, Bin Cao \\
  Zhejiang University of Technology \\
  Hangzhou, China \\\And
  Ruizhe Li \\
  University of Aberdeen \\
  Aberdeen, Scotland, UK \\}

\begin{document}
\maketitle
\begin{abstract}
While Text-to-SQL remains the dominant approach for database interaction, real-world analytics increasingly require the flexibility of general-purpose programming languages such as Python or Pandas to manage file-based data and complex analytical workflows. Despite this growing need, the reliability of Text-to-Python in core data retrieval remains underexplored relative to the mature SQL ecosystem. To address this gap, we introduce BIRD-Python, a benchmark designed for cross-paradigm evaluation. We systematically refined the original dataset to reduce annotation noise and align execution semantics, thereby establishing a consistent and standardized baseline for comparison. Our analysis reveals a fundamental paradigmatic divergence: whereas SQL leverages implicit DBMS behaviors through its declarative structure, Python requires explicit procedural logic, making it highly sensitive to underspecified user intent. To mitigate this challenge, we propose the Logic Completion Framework (LCF), which resolves ambiguity by incorporating latent domain knowledge into the generation process. Experimental results show that (1) performance differences primarily stem from missing domain context rather than inherent limitations in code generation, and (2) when these gaps are addressed, Text-to-Python achieves performance parity with Text-to-SQL. These findings establish Python as a viable foundation for analytical agents—provided that systems effectively ground ambiguous natural language inputs in executable logical specifications. Resources are available at \url{https://github.com/1050727345hu-web/Bird_python}.
\end{abstract}

\section{Introduction}
The rapid advancement of Large Language Models (LLMs) has fundamentally reshaped Natural Language Interface (NLI) research. By allowing users to retrieve information through intuitive natural language queries \citep{qin2022survey,huang2025enable}, NLIs significantly lower technical barriers and enhance the efficiency of human-computer interaction. This paradigm offers substantial potential for broadening data access and analysis across diverse domains. 

Driven by this promise, the research community has focused extensively on Text-to-SQL technology, which translates natural language questions into executable queries for structured databases. This field has developed rapidly, supported by the emergence of comprehensive benchmarks like Spider \citep{yu2018spider} and BIRD \citep{li2024can}, and a progression in techniques from early semantic parsing to modern end-to-end neural generation and schema linking \citep{ruan2023tptu,kong2023tptu,sui2023reboost,zhang2024sgu,cao2024rsl,maamari2024death,safdarian2025schemagraphsql}. However, an exclusive reliance on the Text-to-SQL paradigm presents critical limitations in real-world scenarios. First, vast quantities of practical data reside in standalone files (e.g., CSV, Excel, JSON) rather than managed databases; these formats are not directly queryable via SQL without cumbersome preprocessing. Second, and perhaps more critically, modern analytical needs often require flexible procedural logic and library integration that exceed the expressive capacity of standard declarative SQL, necessitating the computational breadth of general-purpose programming languages.

\begin{figure}[t]
  \includegraphics[width=\linewidth]{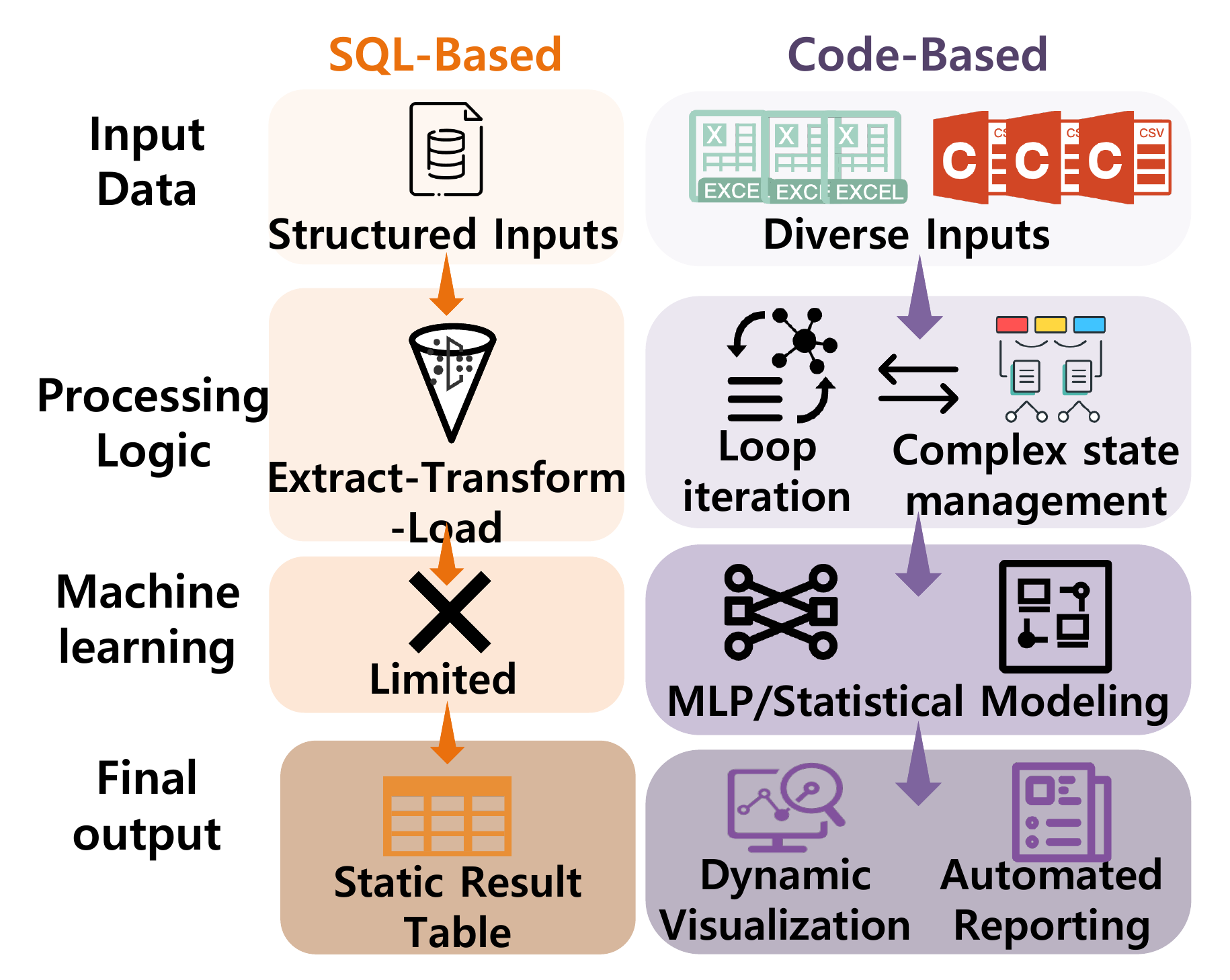}
  \caption{Functional comparison of SQL-based and code-based approaches across key data processing stages. SQL handles structured data and basic queries; code supports these tasks and adds flexibility—direct raw data processing, complex logic, machine learning integration, and automated dynamic output.}
  \label{fig:sqlvscode_biaoge}
\end{figure}

To address these challenges, the Text-to-Python paradigm has emerged as a promising alternative \citep{luo2025tree,jiang2024self,li2025structured}. By generating executable Python code, this approach can directly manipulate file-based data while leveraging powerful libraries like Pandas for robust data manipulation, effectively establishing a foundation for advanced analytics. Consequently, Text-to-Python inherently overcomes the inherent constraints of the Text-to-SQL route, offering a more flexible framework for natural language-driven data exploration (see Figure \ref{fig:sqlvscode_biaoge}).

Despite its theoretical advantages, the Text-to-Python path for data querying and analysis remains largely unexplored. Existing work in code generation predominantly focuses on general programming tasks such as code snippet generation, completion, and repair, as exemplified by benchmarks like HumanEval \citep{chen2021evaluating,zhuo2024bigcodebench} and SWE-Bench \citep{yu2025utboost,dou2024what}. While recent agentic frameworks \citep{spider2,birdinteract,zhou2025rulearena,gu2025structext} and data science benchmarks \citep{jing2024dsbench,egg2025dabstep} have begun to explore broader capabilities, there is a notable absence of a dedicated benchmark to systematically evaluate a model's ability to translate natural language requests into Python code specifically tailored for core data retrieval and analysis. This gap raises a fundamental research question: Can Text-to-Python effectively serve as a viable and more flexible replacement for Text-to-SQL in practical data interaction scenarios?

To investigate this, we conduct a thorough empirical study by adapting the BIRD benchmark \citep{li2024can} to evaluate Text-to-Python performance. To align with real-world scenarios, we transformed the underlying relational databases into standalone file formats (i.e., CSVs) and mapped original SQL-based objectives into equivalent Python execution logic. This process highlights a key paradigmatic distinction: unlike SQL's declarative nature which relies on implicit database rules (e.g., automatic null handling), Python requires explicit, step-by-step procedural logic to define execution behaviors. This fundamental difference exposes the Text-to-Python paradigm to greater challenges regarding ambiguous constraints, necessitating strict definition to ensure precise execution.

Using this refined benchmark, we compare state-of-the-art Text-to-SQL and Text-to-Python models. Our analysis reveals that while Text-to-Python offers comparable performance given sufficient context, its reliance on explicit procedural logic makes it significantly more sensitive to knowledge gaps than SQL. Both paradigms struggle when essential background knowledge is missing \citep{zhang2024benchmarking,liu2024exploring,ouyang2025empirical,zhu2025uncertainty}, but this deficiency is particularly detrimental to Python's need for precise operational steps. This finding suggests that the focus of NLI research should shift toward supplementing models with adequate contextual knowledge. Consequently, we propose the Logic Completion Framework (LCF), which explicitly supplements latent domain knowledge and markedly improves query accuracy across both paradigms.

The contributions of this paper are threefold. (1) First, we propose the first dedicated benchmark BIRD-Python for evaluating Text-to-Python in data querying tasks on file-based environments, enabling a detailed comparative analysis with Text-to-SQL. (2) Second, we uncover the pivotal insight that the need for explicit logic formulation makes knowledge grounding—rather than the choice of paradigm alone—the primary bottleneck in flexible analytics scenarios. (3) Finally, we introduce the LCF, a practical methodology for context completion that explicitly supplements latent domain knowledge and markedly improves the query accuracy of both Text-to-SQL and Text-to-Python systems.




\section{Related Work}
Text-to-SQL approaches have evolved from rule-based matching \citep{zhong2017seq2sql} to advanced systems leveraging LLMs with multi-agent architectures. MAC-SQL \citep{wang2024mac} decomposes queries among specialized agents, while SQL-of-Thought \citep{chaturvedi2025sql} mitigates logical errors through iterative refinement. As the scope of tasks expands beyond traditional database querying to broader data science workflows, Text-to-Python systems have attracted increasing attention \citep{luo2025tree,jiang2024self}. Python code generation offers greater flexibility than SQL but operates without the schema-enforced constraints of databases, which can compromise semantic precision.

Benchmarks have advanced to reflect real-world challenges. Early datasets such as WikiSQL \citep{zhong2017seq2sql} and Spider \citep{yu2018spider} emphasize schema complexity, whereas BIRD \citep{li2024can} incorporates large-scale databases and external knowledge to better represent enterprise-level scenarios. However, evaluating reasoning capabilities remains problematic: execution-based metrics may penalize semantically valid outputs when natural language inputs are ambiguous \citep{pourreza2023evaluating}, indicating that performance limitations often stem from underspecified user intent rather than syntactic inaccuracies.

Contextual ambiguity significantly impacts model performance. Retrieving excessive schema elements introduces noise that degrades accuracy \citep{cao2024rsl,maamari2024death}, while insufficient constraints can lead to invalid logic during error correction attempts \citep{zhang2024benchmarking}. More recent benchmarks like Spider 2.0 \citep{spider2} and BIRD-INTERACT \citep{birdinteract} simulate multi-turn interactions but conflate information retrieval with reasoning processes. In contrast, our approach explicitly separates these components. By aligning SQL and Python execution environments and employing our proposed Logic Completion Framework under controlled conditions, we isolate procedural reasoning from confounding factors such as implicit assumptions and underspecified context.

\begin{figure}[t]
  \includegraphics[width=\linewidth]{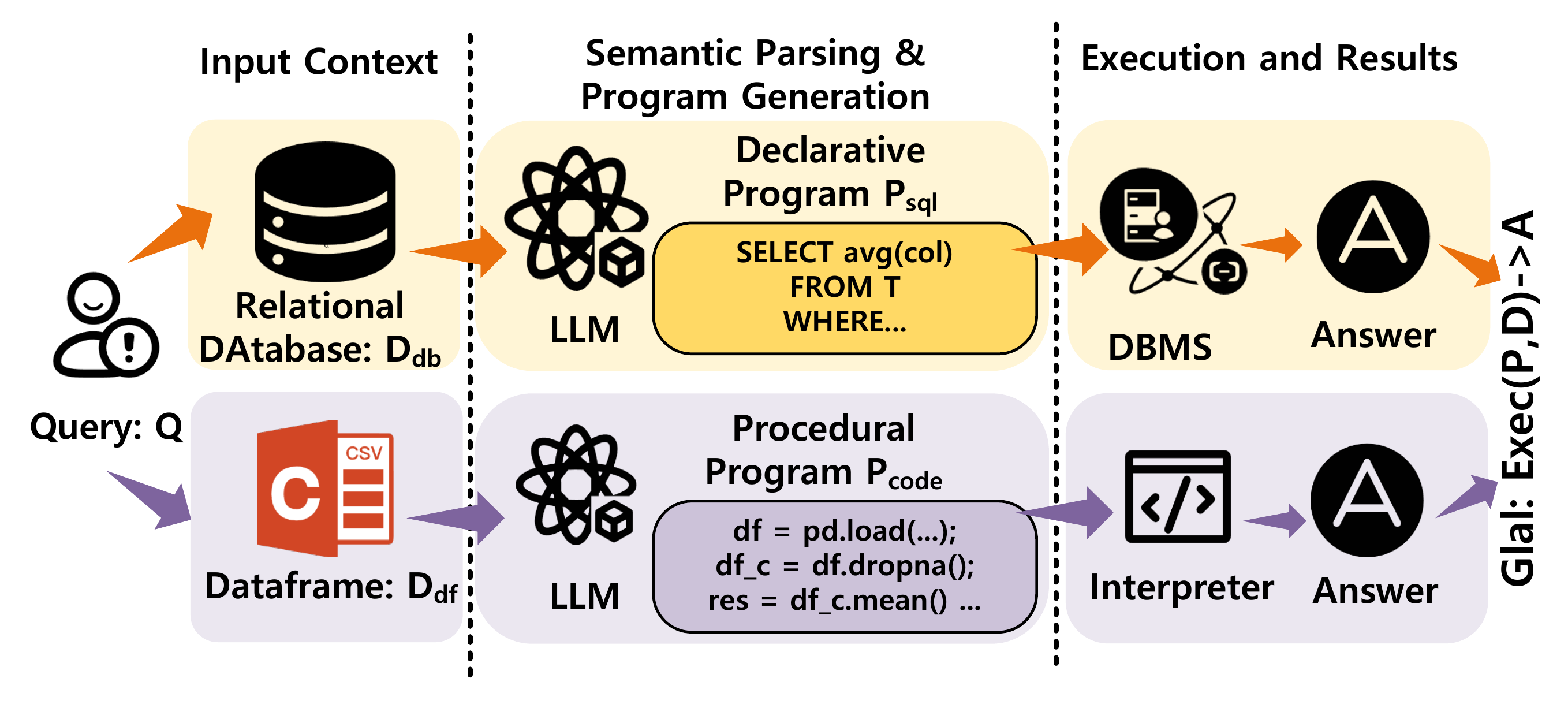}
  \caption{Comparative framework of NLI semantic parsing paradigms. Top: Text-to-SQL maps queries to declarative relational algebra, delegating execution optimization to the DBMS. Bottom: Text-to-Python generates explicit procedural workflows, requiring the model to handle computational reasoning directly.}
  \label{fig:ProblemFormulation}
\end{figure}

\section{Task Formalization and Experimental Framework}
To systematically evaluate whether Text-to-Python can effectively replace Text-to-SQL, we establish a unified experimental framework.  This framework addresses the structural differences between the two programming languages by introducing a standardized task formulation, a corrected dataset to ensure reliability, and consistent evaluation criteria.

\subsection{Problem Formulation}
We unify Text-to-SQL and Text-to-Python as executable program generation tasks (Figure \ref{fig:ProblemFormulation}). Given a query $Q$ and data environment $\mathcal{D}$, the model synthesizes a program $P$ such that $Exec(\mathcal{P}, D) \rightarrow A$. In Text-to-SQL, $\mathcal{D}_{db}$ is a relational database schema $\mathcal{S}$, and the output $P_{sql}$ is a declarative query. Conversely, in Text-to-Python, $\mathcal{D}$ represents in-memory dataframes $\mathcal{D}_{df}$, and the output $P_{code}$ is an imperative script requiring explicit procedural logic.

\subsection{Benchmark Construction}
\label{sec:bird_code_framework}

\subsubsection{Dataset Correction} 

Our study employs the development set of the BIRD benchmark \citep{li2024can}, comprising \textbf{1,534} query-SQL pairs. Each instance is a triplet $(Q, \mathcal{K}, Y_{sql})$ consisting of the natural language question, external knowledge, and the ground truth SQL query. 

However, our preliminary error analysis revealed that evaluation metrics were distorted by annotation noise, where valid model outputs were unfairly penalized due to inaccuracies in the reference SQL. To establish a reliable evaluation foundation for our comparative study, we build upon recent advances in leveraging LLMs for dataset curation \citep{li2024multisql} and the double-blind annotation protocol established in \citep{li2024can}. We implemented a two-stage purification pipeline that integrates Model-in-the-Loop verification with expert review. First, 1,193 consistent queries were automatically verified through consensus among three state-of-the-art LLMs (Qwen3-238B, DeepSeek-R1, and Qwen-Max). The remaining divergent cases underwent double-blind expert review, resulting in 259 corrections for logical inconsistencies or outdated values and the confirmation of 82 original annotations as valid. This rigorous process effectively distinguishes genuine model errors from dataset artifacts.

\begin{figure}[t]
  \includegraphics[width=\linewidth]{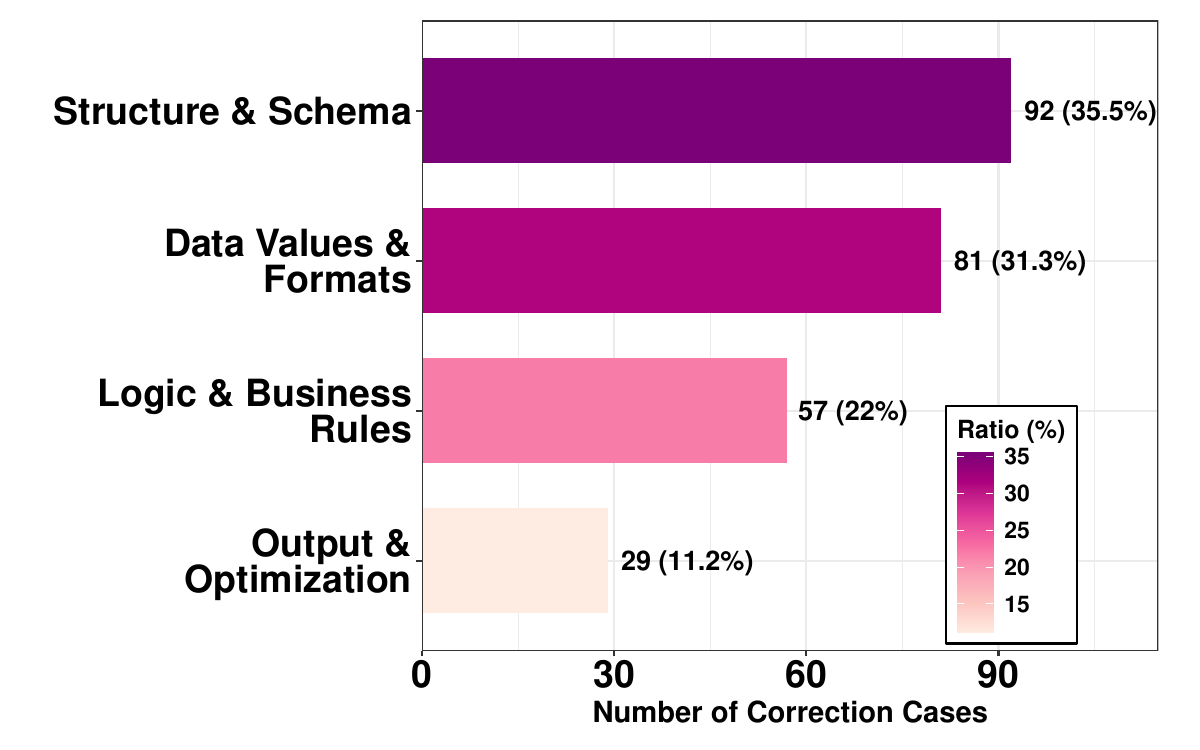}
  \caption{Distribution of difference categories for verified SQL queries.}
  \label{fig:verified_sql_categories}
\end{figure}

We categorized corrections into four dimensions, as illustrated in Figure \ref{fig:verified_sql_categories}. First, we addressed \textit{Structure \& Schema} errors by correcting column hallucinations and missing joins, ensuring the SQL structure faithfully reflects the entity relationships described in natural language. Second, we refined \textit{Data Values \& Formats} by adjusting filter conditions to match actual database values and explicitly handling NULLs, thereby resolving logic gaps caused by dirty data. Third, we performed semantic rectification on \textit{Logic \& Business Rules}, restructuring nested queries and set operations to resolve scope confusion between clauses. Finally, we standardized \textit{Formatting \& Constraints} by adding missing ORDER BY and LIMIT clauses, eliminating non-deterministic outputs to ensure reproducible evaluation.

\begin{figure}[t]
  \includegraphics[width=\linewidth]{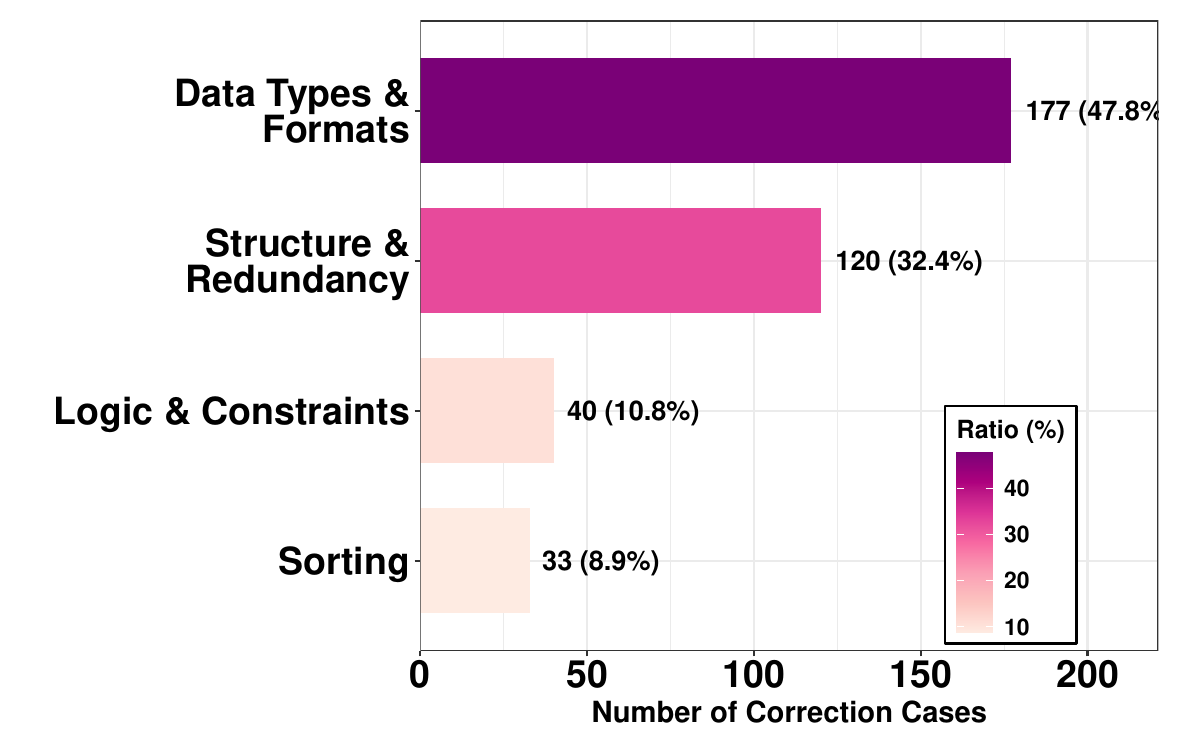}
  \caption{Distribution of difference categories in SQL-to-Python conversion.}
  \label{fig:code_sql_categories}
\end{figure}

\subsubsection{SQL-to-Python Conversion} 
To address the execution paradigm mismatch, we reconstructed SQL logic into Python by explicitly implementing the database's implicit behaviors (e.g., null handling and sort stability). This ensures that the Python ground truth faithfully replicates the SQL execution outcomes within a procedural environment.

\begin{figure*}[t]
  \includegraphics[width=\linewidth]{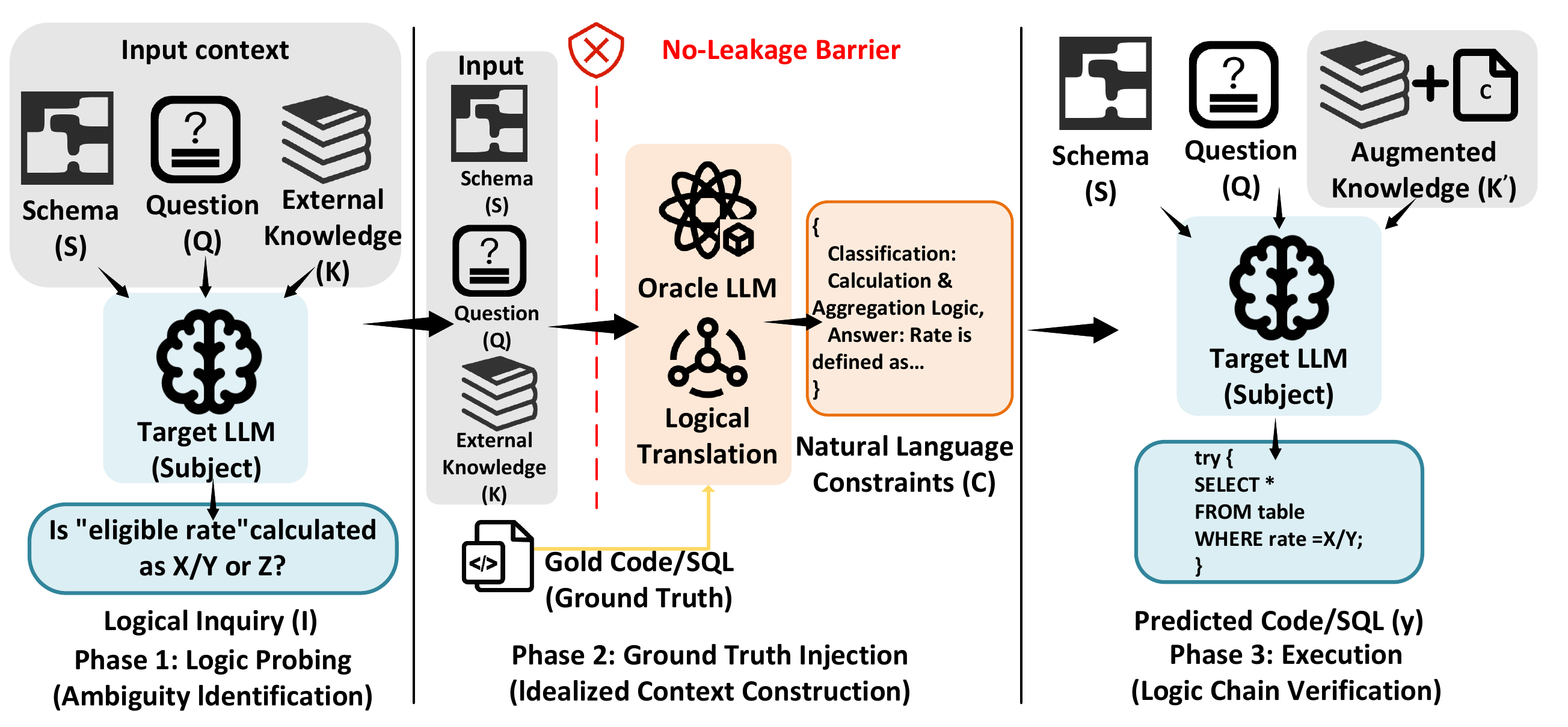}
  \caption{ The Logic Completion Framework (LCF). Standard semantic parsing models the probability of a program $P$ as $P(P | Q, \mathcal{S}, \mathcal{K})$. LCF makes latent domain knowledge explicit by providing logic clarifications ($C_{logic}$), refining the task to $P(P | Q, \mathcal{S}, \mathcal{K} \cup C_{logic})$.}
  \label{fig:LCF}
\end{figure*}

\noindent\textbf{Code Reconstruction}
To establish a valid ground truth for the Text-to-Python task, we reconstructed the SQL logic using Python, ensuring the code explicitly replicated the DBMS-level operations. We employed a similar verification pipeline: after generating code candidates via three SOTA LLMs, we filtered the results to ensure consistency with SQL.

\noindent\textbf{Manual Refinement}
We identified 370 instances requiring manual intervention, which were processed via a double-blind review (see Figure \ref{fig:code_sql_categories}). The primary divergence involves \textit{Data Types \& Formats}; we implemented strict type safety and explicit null semantics to bridge the gap between SQL's permissible implicit casting and Python's strict type safety. Additionally, we addressed \textit{Structure \& Redundancy} by removing redundant structures such as unnecessary joins—often ignored by SQL optimizers but computationally significant in Python. This phase also involved \textit{Logic \& Constraints} refactoring to convert ambiguous assumptions or specific value filters into complete procedural steps that accurately reflect the intent of the user. Finally, we implemented \textit{Sorting} by defining consistent ordering rules for entries with identical values, to ensure the stability of the output results.

\subsubsection{LLM-based Evaluation}
We built an LLM-based evaluation method to evaluate the results of the model execution. Standard metrics such as Execution Accuracy (EX) rely on exact matches between execution outcomes. This approach may produce false negatives in Text-to-Python tasks, where semantically correct outputs are incorrectly classified as erroneous due to superficial discrepancies, such as differences in column ordering or data type representation.

\noindent\textbf{Semantic Validator}
To address this, we implemented an LLM-based semantic validator ($\mathcal{V}_{sem}$) for both SQL and Code contexts. Following the LLM-as-a-Judge methodology \citep{zheng2023judging, kim2024flex, birdinteract}, $\mathcal{V}_{sem}$ determines if the predicted execution result is semantically equivalent to the ground truth, regardless of format.

\noindent\textbf{Human Verification}
To validate the reliability of $\mathcal{V}_{sem}$, we conducted a human audit on a stratified sample of $N=600$ instances. Three domain experts independently assessed each prediction-ground truth pair, with final judgments based on consensus. The manual review confirmed that the automated evaluation results are consistent with expert judgments. This demonstrates that our framework accurately assesses reasoning capability regardless of output format.

\section{Logic Completion Framework}
\label{sec:appendix_lcp}
To address ambiguity in natural language queries, we introduce LCF. Unlike traditional models that generate code directly, LCF first clarifies intent and then generates code. It identifies ambiguous or incomplete requirements and interacts with a domain expert to resolve them. This ensures the final code is based on a complete and logically sound specification.

\subsection{Model Role Design}
The \textit{Subject} (the evaluated model) is responsible for detecting ambiguity (Phase 1) and generating code (Phase 3). Crucially, the Subject generates inquiries independently to test its intrinsic ambiguity detection, rather than relying on a proxy. The \textit{Oracle}, implemented by Qwen3-max, simulates a domain expert. It accesses the ground truth to provide accurate natural language hints ($C_{logic}$) that semantically align with the ground truth of SQL ($H(C_{logic}) \approx H(Y_{gold})$).

\subsection{Dialogue Paradigm}
The interaction follows a three-phase paradigm (Figure \ref{fig:LCF}).

\paragraph{Phase 1: Logic Probing.}
The Subject ($\mathcal{M}_{subject}$) analyzes inputs ($\mathcal{S}, \mathcal{K}, Q$) to identify ambiguity. Instead of immediate code generation, it formulates a clarifying question ($Q_{ambiguity}$) targeting unclear definitions or logic gaps.

\paragraph{Phase 2: Ground Truth Injection.}
The Oracle ($\mathcal{M}_{oracle}$) acts as an expert analyst. Using the inquiry ($Q_{ambiguity}$) and ground truth ($Y_{gold}$), it translates the gold logic into natural language hints ($C_{logic}$). This ensures task requirements are fully specified without leaking code.

\paragraph{Phase 3: Execution.}
The Subject generates the final program ($P_{pred}$) conditioned on the augmented context: $\mathcal{S}, Q, \mathcal{K}, C_{logic}$. This isolates errors caused by reasoning deficiencies from those caused by information deficits. Appendix \ref{sec:appendix_lcp} details the prompts.

We utilized this framework to construct a dataset that addresses information deficits. Rather than relying on human assumptions, we identified missing specifications directly from the model's inquiries in Phase 1.  By incorporating the Oracle's responses, we generated an extra dataset that provides the specific context required for the model's reasoning.

\section{Experiments}

\subsection{Experimental Setup}
We benchmark Text-to-Python paradigm against the traditional Text-to-SQL approach, employing the \textbf{LLM-based Execution Accuracy (EX)} as the primary evaluation metric. Following the standard evaluation protocol of the BIRD benchmark \citep{li2024can}, our input prompts incorporate the database schema, the natural language question, and external knowledge. Specifically for the Text-to-Python setting, we impose additional constraints to ensure the generated code is executable and capable of returning valid results. The detailed prompts are presented in Appendix \ref{sec:appendix_prompts}. We conduct comparative evaluations on both the original BIRD dataset and our refined BIRD-Python dataset to assess the impact of our corrections.

\begin{table*}[t]
  \centering
  \small
  \begin{tabular}{lclcccccccc}
    \hline
    \textbf{Model} & \textbf{Reasoning} & \textbf{Verified} & \multicolumn{4}{c}{\textbf{Code Generation Accuracy (\%)}} & \multicolumn{4}{c}{\textbf{SQL Generation Accuracy (\%)}} \\
     & & & \textbf{Simple} & \textbf{Mod.} & \textbf{Hard} & \textbf{Total} & \textbf{Simple} & \textbf{Mod.} & \textbf{Hard} & \textbf{Total} \\
    \hline
    StarCoder2 & \XSolidBrush & \checkmark & 1.95 & 0.86 & 0.00 & 1.12 & 0.97 & 1.08 & 0.00 & 0.91 \\
    (7B) & & \XSolidBrush & 1.62 & 0.43 & 0.00 & 0.85 & 0.65 & 0.86 & 0.69 & 0.72 \\
    \hline
    AutoCoder & \XSolidBrush & \checkmark & 8.65 & 4.31 & 2.07 & 5.45 & 13.84 & 3.88 & 7.59 & 10.23 \\
    (7B) & & \XSolidBrush & 7.50 & 3.88 & 1.72 & 4.88 & 12.54 & 3.23 & 6.90 & 9.19 \\
    \hline
    TableGPT2 & \XSolidBrush & \checkmark & 25.46 & 17.24 & 13.10 & 19.82 & 40.97 & 25.00 & 18.62 & 34.03 \\
    (7B) & & \XSolidBrush & 23.15 & 15.09 & 11.03 & 17.45 & 39.68 & 22.84 & 18.62 & 32.59 \\
    \hline
    Qwen2.5-Coder & \XSolidBrush & \checkmark & 41.91 & 26.51 & 28.28 & 35.95 & 61.41 & 42.67 & 31.72 & 52.93 \\
    (7B) & & \XSolidBrush & 39.89 & 18.97 & 23.45 & 31.94 & 60.65 & 41.16 & 31.03 & 51.96 \\
    \hline
    Qwen3 & \checkmark & \checkmark & 59.46 & 45.04 & 39.31 & 53.19 & 68.98 & 52.92 & 38.19 & 61.22 \\
    (7B) & & \XSolidBrush & 54.70 & 36.21 & 34.48 & 47.20 & 62.15 & 46.00 & 32.64 & 54.48 \\
    \hline
    Qwen3 & \checkmark & \checkmark & 54.70 & 37.93 & 37.24 & 47.98 & 65.84 & 48.28 & 42.76 & 58.34 \\
    (14B) & & \XSolidBrush & 65.84 & 32.33 & 29.66 & 42.89 & 59.46 & 44.18 & 39.31 & 52.93 \\
    \hline
    Qwen2.5-Coder & \XSolidBrush & \checkmark & 60.76 & 45.69 & 40.69 & 54.30 & 63.78 & 51.29 & 50.34 & 58.74 \\
    (14B) & & \XSolidBrush & 51.24 & 33.84 & 35.86 & 44.52 & 63.24 & 46.55 & 44.14 & 56.39 \\
    \hline
    Qwen2.5-Coder & \XSolidBrush & \checkmark & 62.16 & 49.78 & 42.76 & 56.58 & 66.70 & 50.65 & 41.38 & 59.45 \\
    (32B) & & \XSolidBrush & 58.38 & 41.81 & 37.93 & 51.43 & 64.11 & 47.84 & 39.31 & 56.84 \\
    \hline
    Qwen3-Coder & \XSolidBrush & \checkmark & 49.51 & 41.59 & 35.86 & 45.83 & 67.35 & 55.82 & 43.45 & 61.60 \\
    (30B) & & \XSolidBrush & 46.92 & 34.05 & 32.41 & 41.53 & 64.11 & 51.29 & 43.54 & 58.28  \\
    \hline
    Qwen3 & \checkmark & \checkmark & 65.33 & 51.94 & 54.48 & 60.10 & 69.08 & 54.09 & 40.00 & 61.80 \\
    (32B) & & \XSolidBrush & 60.22 & 42.24 & 45.52 & 53.32 & 61.84 & 47.84 & 35.17 & 55.08 \\
    \hline
    Qwen-Max & \XSolidBrush & \checkmark & 65.73 & 53.66 & 52.41 & 60.82 & 68.32 & 51.72 & 40.00 & 59.20 \\
     & & \XSolidBrush & 59.57 & 45.04 & 42.76 & 53.59 & 63.14 & 45.47 & 34.48 & 55.08 \\
    \hline
    Qwen3-Coder-Plus & \XSolidBrush & \checkmark & 65.95 & 55.82 & 53.10 & 61.60 & 65.30 & \textbf{56.25} & 50.34 & 61.15 \\
     & & \XSolidBrush & 60.32 & 45.26 & 42.07 & 53.98 & 62.92 & 53.45 & 50.34 & 58.87 \\
    \hline
    DeepSeek-R1 & \checkmark & \checkmark & \textbf{69.95} & 53.02 & 45.52 & 62.52 & 67.54 & 53.48 & \textbf{57.93} & 62.32 \\
     & & \XSolidBrush & 62.38 & 45.04 & 48.28 & 55.74 & 62.81 & 44.40 & 40.00 & 55.08 \\
    \hline
    Qwen3-Max & \XSolidBrush & \checkmark & 67.89 & \textbf{59.91} & 46.21 & \textbf{63.43} & \textbf{69.41} & 55.82 & 46.21 & \textbf{63.10} \\
     & & \XSolidBrush & 62.27 & 43.75 & 47.59 & 55.28 & 65.51 & 52.59 & 52.41 & 60.37 \\
    \hline
  \end{tabular}
  \caption{Experimental results on the BIRD-Python benchmark (Python) and original BIRD benchmark (SQL). We report Execution Accuracy (EX) across Simple, Moderate (Mod.), and Challenging (Hard) subsets. "\checkmark" denotes our verified dataset; "\XSolidBrush" denotes the original BIRD dev set.}
  \label{tab:main_results}
\end{table*}

\subsection{Baselines}
To ensure a comprehensive evaluation, we select a diverse range of open-source and closed-source LLMs. All open-source models are deployed locally on a server equipped with six NVIDIA A6000 GPUs. We utilize default inference parameters, with a temperature of $0.7$ and a top-$p$ of $0.95$. Our baseline selection includes: (1) \textbf{StarCoder2}~\cite{lozhkov2024starcoder2} ($7$B), \textbf{AutoCoder}~\cite{lei2024autocoder} ($7$B), and \textbf{TableGPT2}~\cite{su2024tablegpt2} ($7$B); (2) The extensive \textbf{Qwen} family, covering the \textbf{Qwen-Coder-2.5} series~\cite{hui2024qwen25coder}, the \textbf{Qwen3} series~\cite{yang2025qwen3} (including both reasoning and coder variants), and the latest closed-source \textbf{Qwen3-Max}; and (3) The reasoning-oriented \textbf{DeepSeek-R1}~\cite{guo2025deepseekr1}.

\subsection{Main Results}
Table \ref{tab:main_results} presents the performance of baseline models. First, we observe a performance gap between the two paradigms. For smaller models, Python generation accuracy is lower than SQL; for instance, Qwen2.5-Coder ($7$B) decreases from $52.93\%$ on SQL to $35.95\%$ on Python. This performance drop empirically validates our hypothesis that the requirement for explicit procedural definition in Python creates a higher reasoning barrier for smaller models compared to the declarative nature of SQL.

However, this gap decreases with larger or reasoning-enhanced models. The Qwen3 series and DeepSeek-R1 show smaller performance drops between SQL and Python tasks. Qwen3-Max achieves $63.43\%$ accuracy in Python generation, slightly higher than DeepSeek-R1 ($62.52\%$). In contrast, models specialized for specific tasks show lower transferability. While AutoCoder and TableGPT2 achieve scores comparable to other baselines on SQL, their performance on Python tasks is lower ($5.45\%$ and $19.82\%$, respectively), suggesting that SQL-specific or general code-completion training may not directly translate to the Pandas-based data analysis task.

Finally, comparing the Verified and Original settings indicates the impact of data quality. Models such as Qwen3-Max and DeepSeek-R1 show higher accuracy on the Verified benchmark ($+8.15\%$ and $+6.78\%$, respectively). This increase suggests that the original dataset contained noise that affected evaluation scores, while the verified set provides a more accurate assessment of model outputs.

\begin{figure}[t]
  \includegraphics[width=\linewidth]{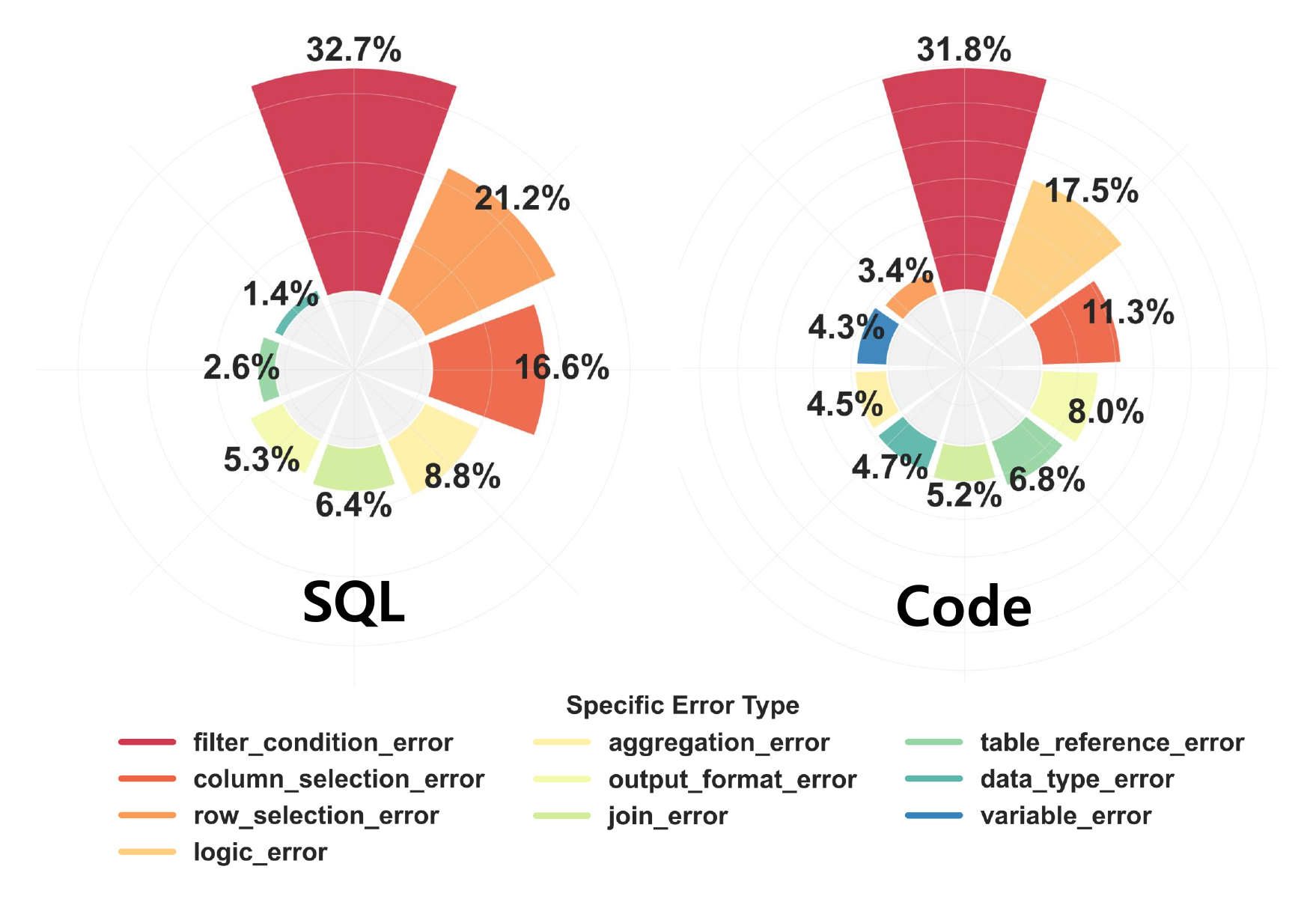}
  \caption{Four Dimensions of Semantic Mismatch causing Execution Errors.}
  \label{fig:ExecutionErrors}
\end{figure}

\begin{table*}[t]
  \centering
  \small
  \begin{tabular}{llcccccccc}
    \hline
    \multirow{2}{*}{\textbf{Model}} & \multirow{2}{*}{\textbf{Setting}} & \multicolumn{4}{c}{\textbf{Code Generation Accuracy (\%)}} & \multicolumn{4}{c}{\textbf{SQL Generation Accuracy (\%)}} \\
    \cline{3-10}
     & & \textbf{Simple} & \textbf{Mod.} & \textbf{Hard} & \textbf{Total} & \textbf{Simple} & \textbf{Mod.} & \textbf{Hard} & \textbf{Total} \\
    \hline
    \textbf{Qwen3-7B} & LCF & \textbf{79.46} & \textbf{61.64} & \textbf{48.97} & \textbf{71.19} & \textbf{80.65} & \textbf{60.99} & \textbf{45.52} & \textbf{71.38} \\
    (7B) & Standard & 59.46 & 45.04 & 39.31 & 53.19 & 68.98 & 52.92 & 38.19 & 61.22 \\
    \hline
    \textbf{Qwen3-32B} & LCF & \textbf{78.16} & \textbf{63.15} & \textbf{66.21} & \textbf{72.49} & \textbf{80.54} & \textbf{62.93} & \textbf{54.48} & \textbf{72.75} \\
    (32B) & Standard & 65.33 & 51.94 & 54.48 & 60.10 & 69.08 & 54.09 & 40.00 & 61.80 \\
    \hline
    \textbf{Qwen3-Max} & LCF & \textbf{77.84} & \textbf{66.16} & \textbf{66.90} & \textbf{73.21} & \textbf{85.95} & \textbf{72.63} & \textbf{66.21} & \textbf{80.05} \\
    ($\sim$1T) & Standard & 67.89 & 59.91 & 46.21 & 63.43 & 69.41 & 55.82 & 46.21 & 63.10 \\
    \hline
  \end{tabular}
  \caption{Ablation study of the LCF. We evaluate performance on the sanitized BIRD-Python dataset in two settings: "Standard" (baseline without LCF) and "LCF" (with logical clarifications). The highest score for each model is bolded.}
  \label{tab:LCF_results}
\end{table*}

\subsubsection{Error Analyze}
We analyze execution failures using the error categories defined in our framework. Figure \ref{fig:ExecutionErrors} shows two trends: shared difficulties in natural language understanding and paradigm-specific error patterns.

Misunderstanding the question's intent constitutes the primary source of error. We observe that \textit{Filter Condition Errors} are the most prevalent in both SQL (32.7\%) and Code (31.7\%). This consistency indicates that natural language understanding plays a critical role in both paradigms. Models frequently misinterpret ambiguous user intents—such as vague temporal references or implicit numerical thresholds—translating them into incorrect logical predicates. This implies that, in the absence of explicit clarification, models in both settings tend to resolve linguistic ambiguity incorrectly by mapping it onto schema elements.

Beyond this common challenge, error types differ by paradigm. In SQL, errors frequently involve \textit{Row Selection} (21.2\%) and \textit{Column Selection} (16.6\%), often related to clauses like \texttt{LIMIT} or \texttt{DISTINCT}. In contrast, Python generation exhibits a higher rate of \textit{Logic Errors} (17.5\%) compared to SQL (0.3\%). This reflects the difference in execution: SQL engines manage execution plans implicitly, whereas Python requires explicit implementation of data manipulation steps, increasing the likelihood of procedural errors.

These results illustrate the tension between underspecified user intent and the precision required for program execution. While SQL abstracts away procedural details, Python's requirement for explicit logic definition tends to increase the risk of hallucinated specifications. This suggests that current evaluations may conflate ambiguity resolution with code generation capability, highlighting the need to distinguish missing specifications from logical reasoning ability.

\subsection{LCF Test}
To separate reasoning capability from information deficits, we apply the LCF to the verified BIRD-Python dataset. We select the Qwen3 family (7B, 32B, Max), which shows strong performance across scales, to evaluate gains when latent specifications are resolved. LCF enriches queries with explicit specifications (see Section \ref{sec:appendix_lcp}), isolating execution failures caused by underspecified requirements.

\begin{figure}[t]
  \includegraphics[width=\linewidth]{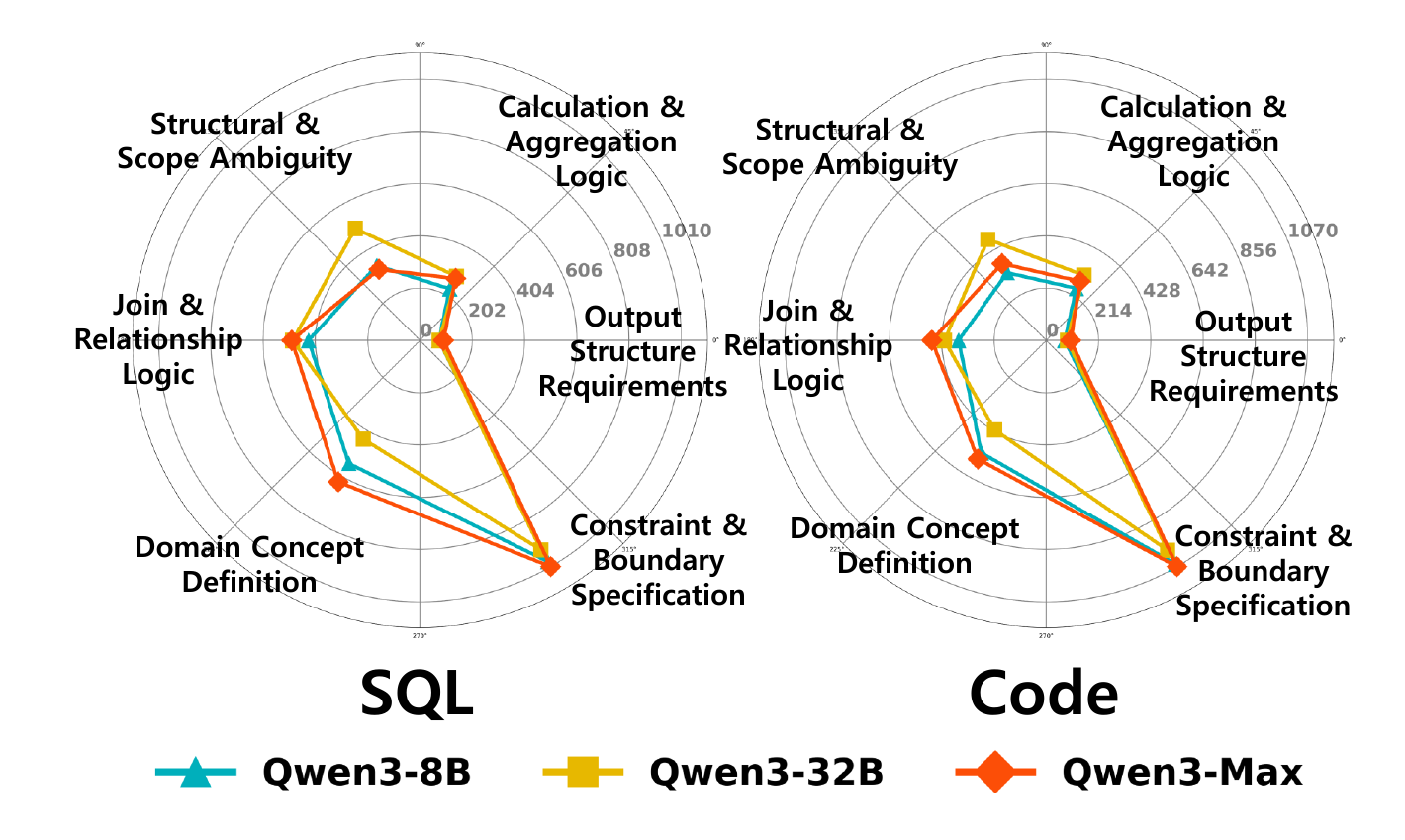}
  \caption{Distribution of inquiry categories for SQL and Code generation across Qwen3 model scales.}
  \label{fig:Characterizing Epistemic Ambiguity}
\end{figure}

\subsubsection{Ablation Study Results}
Table \ref{tab:LCF_results} compares model performance under Standard and LCF settings. Resolving ambiguity improves accuracy across all models. For Qwen3-7B, performance increases in both Code ($53.19\% \to 71.19\%$) and SQL ($61.22\% \to 71.38\%$). This suggests that incomplete inputs contribute to performance limitations in the Standard setting.

Additionally, the performance gap between SQL and Code generation decreases under well-specified clarifications. With LCF, Qwen3-32B achieves similar accuracy in Code ($72.49\%$) and SQL ($72.75\%$), indicating that Python-based generation performs comparably to SQL when semantic specifications are clear.

Finally, LCF clarifies the impact of model scaling. In the Standard setting, ambiguity appears to obscure performance differences across model sizes. Under LCF, larger models show clearer advantages: Qwen3-Max outperforms Qwen3-7B by approximately 9 percentage points in SQL ($80.05\%$ vs. $71.38\%$). This demonstrates that larger models more effectively leverage explicit specifications.

\subsubsection{Analysis of Clarification Inquiries}
We categorize the clarification inquiries from Phase 1 to identify sources of ambiguity (Figure \ref{fig:Characterizing Epistemic Ambiguity}). The distribution of inquiry types remains consistent across model sizes. "Constraint \& Boundary Specification" is the dominant category, whereas inquiries regarding "Output Structure Requirements" are minimal. This suggests that models generally handle syntactic requirements effectively but have difficulty resolving ambiguous business logic without explicit context. Notably, this pattern persists across scales: larger models like Qwen3-Max identify logical gaps more precisely rather than attempting implicit resolution. This suggests that model scaling alone may not fully resolve domain-specific ambiguity.

\section{Conclusion}
This paper evaluates the Text-to-Python paradigm as a viable and flexible alternative to Text-to-SQL for data interaction. Using the BIRD-Python benchmark, we demonstrate that Python can effectively replicate SQL's core data retrieval capabilities in file-based environments. However, the transition from SQL’s declarative syntax to Python’s procedural formulation increases sensitivity to underspecified user intent. While Python enables greater flexibility for advanced analytics, this benefit depends on precise specification of execution logic. Through application of the LCF, we identify the primary performance bottleneck not in code generation itself, but in gaps in latent domain knowledge. Therefore, Text-to-Python can serve as a robust alternative to Text-to-SQL—provided that systems are capable of translating ambiguous natural language inputs into logically precise, executable code.

\section*{Limitations}

Although BIRD-Python and LCF provide a controlled framework for cross-paradigm evaluation, limitations exist. First, we provided explicit DDL schemas to Text-to-Python models to ensure fair comparison. While this isolates procedural reasoning, it may yield higher performance than scenarios involving raw, schema-less files.

Second, LCF utilizes an Oracle with access to ground truth. While useful for diagnosis, this reliance assumes high-quality annotations; practical deployments would likely depend on human feedback, introducing latency not modeled here.

Third, the evaluation focuses on logical complexity using clean data. Consequently, it does not fully address issues related to data quality, such as type inference noise common in production. Future work could examine the interaction between data quality and intent ambiguity.


\bibliography{custom}

\appendix
\section{Baseline Prompt Specifications}
\label{sec:appendix_prompts}

This appendix provides the prompt templates for the baseline Text-to-SQL and Text-to-Python generation tasks (Phase 3 in the LCF framework utilizes the Text-to-Python template). Both paradigms receive identical schema information \citep{li2024can}.

\subsection{Text-to-SQL Prompt}
We follow the standard BIRD benchmark protocol. The system instruction and prompt template are defined in Table \ref{tab:sql_system} and Table \ref{tab:sql_prompt}, respectively.

\begin{table}[h]
    \small
    \centering
    \begin{tabular}{p{0.95\linewidth}}
    \toprule
    \textbf{Text-to-SQL System Instruction} \\
    \midrule
    You are a SQL assistant. Only return the SQL query without any explanation. \\
    \bottomrule
    \end{tabular}
    \caption{System instruction for Text-to-SQL generation.}
    \label{tab:sql_system}
\end{table}

\begin{table}[t!]
    \small
    \centering
    \begin{tabular}{p{0.95\linewidth}}
    \toprule
    \textbf{Text-to-SQL Prompt Template} \\
    \midrule
    /* [Schema Info: DDL] */ \\
    CREATE TABLE `table\_name` (...); \\
    /* \\
    ... \\
     \\
    -- External Knowledge: \{knowledge\} \\
    -- Using valid SQLite and understanding External Knowledge, \\
    -- answer the following questions for the tables provided above. \\
    -- \{question\} \\
    SELECT \\
    \bottomrule
    \end{tabular}
    \caption{Complete Text-to-SQL prompt template including schema information.}
    \label{tab:sql_prompt}
\end{table}

\subsection{Text-to-Python Prompt}
To evaluate procedural reasoning, we require a Pandas-based generation format. The prompt excludes SQL syntax and requires the use of \texttt{pd.read\_csv} to simulate file-based workflows. The system instruction and prompt template are provided in Table \ref{tab:code_system} and Table \ref{tab:code_prompt}.

\begin{table}[t!]
    \small
    \centering
    \begin{tabular}{p{0.95\linewidth}}
    \toprule
    \textbf{Text-to-Python System Instruction} \\
    \midrule
    You are an expert Python code generator specializing in pandas data analysis. \\
    Return runnable Python code only. No explanations or markdown. \\
    Strictly use 'import pandas as pd' and pd.read\_csv('<table>.csv'). \\
    Do NOT use or mention SQL/SELECT/JOIN/CREATE/WHERE/etc. \\
    Do NOT define functions or classes (no 'def', 'lambda', 'class'). \\
    \bottomrule
    \end{tabular}
    \caption{System instruction for Text-to-Python generation.}
    \label{tab:code_system}
\end{table}

\begin{table}[t!]
    \small
    \centering
    \begin{tabular}{p{0.95\linewidth}}
    \toprule
    \textbf{Text-to-Python Prompt Template} \\
    \midrule
    /* [Schema Info: Same as Text-to-SQL] */ \\
    ... \\
     \\
    \# Task Description: \\
    \# Generate runnable pandas code only. No explanations, \\
    \# no markdown, no JSON. \\
    \# Requirements: \\
    \# 1) Use: import pandas as pd \\
    \# 2) Read tables strictly via pd.read\_csv('<table>.csv') \\
    \# 3) Do NOT use or mention SQL/ SELECT/ JOIN/ CREATE/ WHERE/ etc. \\
    \# 4) Do NOT define functions or classes (no 'def', 'lambda', 'class') \\
    \# 5) Prefer clear variable names; keep code executable end-to-end \\
    \# 6) Use result to record the final result, and finally \\
    \#    print(result) to print the final result. \\
     \\
    \# External Knowledge: \{knowledge\} \\
    \# Question: \{question\} \\
    CODE \\
    \bottomrule
    \end{tabular}
    \caption{Complete Text-to-Python prompt template with pandas-specific requirements.}
    \label{tab:code_prompt}
\end{table}

\section{Rectification Case Studies}
\label{sec:appendix_rectification}
We present case studies for the four error categories defined in Section \ref{sec:bird_code_framework}. These examples illustrate logical inconsistencies in the original BIRD dataset and the corrections implemented for BIRD-Python.

\subsection{Category 1: Schema Linking \& Structural Alignment}
The "Schema Linking \& Structural Alignment" category accounts for 35.5\% of errors. It involves cases where the SQL structure does not match the logic required by the query, particularly regarding join semantics (e.g., \texttt{INNER} vs. \texttt{LEFT} joins) and aggregation scope.

Table \ref{tab:rectification_example_structural} shows a representative instance. The original annotation used an \texttt{INNER JOIN} combined with a \texttt{WHERE} clause. This structure removed cities that satisfied the primary condition (K-8 Magnet schools) but lacked the secondary attribute (Multiple Provision Types), and failed to include the "Total Schools" column. The rectified SQL uses a \texttt{LEFT JOIN} and conditional aggregation (\texttt{CASE WHEN}), retaining all relevant cities and calculating the required statistics.

\begin{table}[t!]
    \small
    \centering
    \begin{tabular}{p{0.95\linewidth}}
    \toprule
    \textbf{Case Study: Structural Alignment \& Join Semantics} \\
    \midrule
    \textbf{Natural Language Query:} \\
    Of the schools that offers a magnet program serving a grade span of Kindergarten to 8th grade, how many offers Multiple Provision Types? List the number of cities that offers a Kindergarten to 8th grade span and indicate how many schools are there serving such grade span for each city. \\
    \midrule
    \textbf{Original SQL:} \\
    \texttt{SELECT T2.City, COUNT(T2.CDSCode) FROM frpm AS T1 INNER JOIN schools AS T2 ON T1.CDSCode = T2.CDSCode WHERE T2.Magnet = 1 AND T2.GSoffered = 'K-8' AND T1.`NSLP Provision Status` = 'Multiple Provision Types' GROUP BY T2.City} \\
    \textit{Defect:} The use of \texttt{INNER JOIN} with the \texttt{WHERE} filter restricts the result set to cities containing schools with "Multiple Provision Types", excluding cities that satisfy the K-8 Magnet condition but lack this specific provision. Additionally, the query omits the calculation for the total school count per city. \\
    \midrule
    \textbf{Rectified SQL (Gold):} \\
    \texttt{SELECT T2.City, COUNT(DISTINCT CASE WHEN T1.`NSLP Provision Status` = 'Multiple Provision Types' THEN T2.CDSCode END), COUNT(DISTINCT T2.CDSCode) FROM schools AS T2 LEFT JOIN frpm AS T1 ON T2.CDSCode = T1.CDSCode WHERE T2.GSoffered = 'K-8' AND T2.Magnet = 1 GROUP BY T2.City} \\
    \textit{Correction:} Uses \texttt{LEFT JOIN} to retain all cities meeting the primary K-8 Magnet criteria. Applies conditional aggregation to calculate both the subset (Multiple Provision) and the total school count, addressing the dual-quantification requirement of the prompt. \\
    \bottomrule
    \end{tabular}
    \caption{Comparison of Original and Rectified SQL for a complex aggregation task. The rectification resolves structural misalignment caused by incorrect filtering and join selection.}
    \label{tab:rectification_example_structural}
\end{table}

\subsection{Category 2: Data Consistency \& Integrity}
The "Data Consistency \& Integrity" category (31.3\%) involves cases where the SQL logic incorrectly maps natural language entities to database values. Common issues include incorrect literals, string matching errors, or format mismatches.

Table \ref{tab:rectification_example_data} presents an example of a value mismatch. The user queries for "Alameda" county, but the original SQL filters for "Lake", a value not present in the query. This error likely stems from incorrect entity mapping. The rectification process aligns the SQL literal with the information provided in the natural language hints.

\begin{table}[t!]
    \small
    \centering
    \begin{tabular}{p{0.95\linewidth}}
    \toprule
    \textbf{Case Study: Data Consistency \& Value Alignment} \\
    \midrule
    \textbf{Natural Language Query:} \\
    How many schools in merged Alameda have number of test takers less than 100? \\
    \midrule
    \textbf{Original SQL:} \\
    \texttt{SELECT COUNT(T1.CDSCode) FROM schools AS T1 INNER JOIN satscores AS T2 ON T1.CDSCode = T2.cds WHERE T1.StatusType = 'Merged' AND T2.NumTstTakr < 100 AND T1.County = 'Lake'} \\
    \textit{Defect:} The query filters by \texttt{'Lake'} instead of \texttt{'Alameda'}, using a literal value that does not appear in the prompt. \\
    \midrule
    \textbf{Rectified SQL (Gold):} \\
    \texttt{SELECT COUNT(T1.CDSCode) FROM schools AS T1 INNER JOIN satscores AS T2 ON T1.CDSCode = T2.cds WHERE T1.StatusType = 'Merged' AND T2.NumTstTakr < 100 AND T1.County = 'Alameda'} \\
    \textit{Correction:} Maps the token "Alameda" to the corresponding database value \texttt{'Alameda'}, ensuring the query filters the correct data partition. \\
    \bottomrule
    \end{tabular}
    \caption{Comparison of Original and Rectified SQL demonstrating a Data Consistency error. The rectification resolves the mismatch between the user's constraint and the SQL literal.}
    \label{tab:rectification_example_data}
\end{table}

\subsection{Category 3: Semantic Logic \& Business Rules}
The "Semantic Logic \& Business Rules" category ($22.0\%$) covers errors where the SQL implementation does not match the query's business intent, particularly regarding conditional logic and scope. These errors often involve incorrect translation of natural language quantifiers (e.g., "all", "only") into SQL predicates like \texttt{NOT EXISTS} or \texttt{HAVING}.

Table \ref{tab:rectification_example_semantics} shows a "Quantification Scope" error. The user queries for "single bond molecules," implying that \textit{all} bonds in the molecule must be of the single type. However, the original SQL uses a \texttt{JOIN} with a filter, which functions as an existential check (selecting molecules with \textit{at least one} single bond). This logic incorrectly includes molecules with mixed bond types. The rectification uses \texttt{NOT EXISTS} to exclude molecules containing non-single bonds, satisfying the universal quantifier.

\begin{table}[t!]
    \small
    \centering
    \begin{tabular}{p{0.95\linewidth}}
    \toprule
    \textbf{Case Study: Semantic Logic \& Quantification Scope} \\
    \midrule
    \textbf{Natural Language Query:} \\
    Among the single bond molecule id, which molecules are not carcinogenic? \\
    \midrule
    \textbf{Original SQL:} \\
    \texttt{SELECT DISTINCT T1.molecule\_id FROM bond AS T1 INNER JOIN molecule AS T2 ON T1.molecule\_id = T2.molecule\_id WHERE T2.label = '-' AND T1.bond\_type = '-'} \\
    \textit{Defect:} The query uses an existential filter (\texttt{bond\_type = '-'}), selecting molecules with \textit{at least one} single bond instead of those composed \textit{exclusively} of single bonds. \\
    \midrule
    \textbf{Rectified SQL (Gold):} \\
    \texttt{SELECT T2.molecule\_id FROM molecule T2 WHERE T2.label = '-' AND NOT EXISTS (SELECT 1 FROM bond WHERE molecule\_id = T2.molecule\_id AND bond\_type != '-')} \\
    \textit{Correction:} Uses \texttt{NOT EXISTS} to ensure no non-single bonds exist for the molecule, matching the definition of a "single bond molecule." \\
    \bottomrule
    \end{tabular}
    \caption{Comparison of Original and Rectified SQL illustrating a Semantic Logic error. The rectification fixes the quantification scope from existential ("any") to universal ("all") using negative existential logic.}
    \label{tab:rectification_example_semantics}
\end{table}

\subsection{Category 4: Formatting \& Operational Constraints}
The "Formatting \& Operational Constraints" category ($11.2\%$) covers cases where the SQL logic is correct but the output formatting does not match requirements, such as column ordering or tie-breaking. These issues can lead to false negatives in automated evaluation.

Table \ref{tab:rectification_example_formatting} shows a "Column Ordering" error. The user requests the address components in the sequence: "Street, City, Zip and State". The original SQL retrieves the correct fields but swaps the last two, returning "State" before "Zip". Although semantically minor, this discrepancy affects evaluation accuracy. The rectification reorders the projection clause to match the requested sequence.

\begin{table}[t!]
    \small
    \centering
    \begin{tabular}{p{0.95\linewidth}}
    \toprule
    \textbf{Case Study: Formatting \& Output Determinism} \\
    \midrule
    \textbf{Natural Language Query:} \\
    What is the complete address of the school with the lowest excellence rate? Indicate the Street, City, Zip and State. \\
    \midrule
    \textbf{Original SQL:} \\
    \texttt{SELECT T2.Street, T2.City, T2.State, T2.Zip FROM satscores AS T1 INNER JOIN schools AS T2 ON T1.cds = T2.CDSCode ORDER BY CAST(T1.NumGE1500 AS REAL) / T1.NumTstTakr ASC LIMIT 1} \\
    \textit{Defect:} The query projects columns in the order \texttt{(Street, City, State, Zip)}, inconsistent with the user instruction to present \texttt{Zip} before \texttt{State}. \\
    \midrule
    \textbf{Rectified SQL (Gold):} \\
    \texttt{SELECT T2.Street, T2.City, T2.Zip, T2.State FROM satscores AS T1 INNER JOIN schools AS T2 ON T1.cds = T2.CDSCode ORDER BY CAST(T1.NumGE1500 AS REAL) / T1.NumTstTakr ASC LIMIT 1} \\
    \textit{Correction:} Reorders the \texttt{SELECT} clause to match the requested output format \texttt{(Street, City, Zip, State)}. \\
    \bottomrule
    \end{tabular}
    \caption{Comparison of Original and Rectified SQL demonstrating a Formatting error. The rectification fixes the column permutation to comply with the output order requirement.}
    \label{tab:rectification_example_formatting}
\end{table}

\section{Structure-Agnostic Validator Specifications}
\label{sec:appendix_validator}

Table \ref{tab:validator_prompt} provides the specifications for the Structure-Agnostic Validator ($\mathcal{V}_{sem}$). The prompt is designed to verify semantic equivalence, ignoring format variations to focus on content accuracy.

\begin{table}[t!]
    \small
    \centering
    \begin{tabular}{p{0.95\linewidth}}
    \toprule
    \textbf{Structure-Agnostic Validator Instruction} \\
    \midrule
    \textbf{System Role:} \\
    You are a professional output-equivalence validator. Decide whether Output 1 and Output 2 are semantically and materially equivalent. Follow these rules strictly: \\
    \vspace{0.3em}
    \textbf{Core Equivalence Rules:} \\
    1. \textbf{Content Over Format:} Ignore data presentation (e.g., CSV vs. JSON, whitespace) and metadata (e.g., Pandas Index). Focus solely on the actual content. \\
    2. \textbf{Structural Invariance:} Flatten nested structures. A scalar, a single-element list, and a 1xN array are equivalent if they contain the same value (e.g., ['apple'] == 'apple'). \\
    3. \textbf{Order Insensitivity:} Unless explicit sorting is required, treat lists and table rows as multisets. Order does not matter. \\
    4. \textbf{Type Normalization:} Normalize numeric types (1.0 == 1), booleans (True/yes/1), and date formats before comparison. \\
    5. \textbf{Superset Validity:} If one output contains extra descriptive columns (labels) and the other is value-only, they are equivalent if the value projection matches exactly. \\
    6. \textbf{Numeric Tolerance:} Percentages and fractions are equivalent (e.g., 0.227 == 22.7\%) within standard rounding tolerance. \\
    7. \textbf{De-duplication:} If one output contains duplicates and the other is unique, compare the sets of unique values. \\
    \vspace{0.3em}
    \textbf{Evaluation Task:} \\
    Output 1: \{predicted\_output\} \\
    Output 2: \{ground\_truth\} \\
    Respond with exactly one word: "Correct" if equivalent, "Incorrect" otherwise. \\
    \bottomrule
    \end{tabular}
    \caption{Prompt specifications for the Structure-Agnostic Validator ($\mathcal{V}_{sem}$). The system applies normalization rules (summarized above) to distinguish semantic correctness from format variations.}
    \label{tab:validator_prompt}
\end{table}

\section{Detailed Case Studies of Semantic Mismatch}
\label{sec:appendix_cases}

This appendix presents case studies supporting the discrepancy taxonomy in the main text. We analyze examples across three dimensions—Null Semantics, Operational Determinism, and Granularity Mismatch—to demonstrate execution divergence. These cases illustrate the mechanisms behind model failures and link the concept of impedance mismatch to empirical results.

Table \ref{tab:mismatch_example} illustrates the mismatch in null-value handling. The SQL standard \texttt{GROUP BY} clause preserves \texttt{NULL} keys, whereas Pandas \texttt{groupby} defaults to \texttt{dropna=True}, removing rows with \texttt{NaN} keys. The model failed to set \texttt{dropna=False}, causing valid records to be discarded.

Table \ref{tab:tie_breaking_mismatch} demonstrates discrepancies caused by nondeterministic sorting. With identical values at the limit boundary ($N=5$) and no secondary sort key, SQL and Pandas rely on different default behaviors. This leads to inconsistencies in the result set.

Table \ref{tab:granularity_mismatch} highlights granularity mismatches driven by constraint inference. The term "district" led the model to apply a filter (\texttt{rtype='D'}), assuming district-level aggregation. However, the ground truth SQL selects the global maximum without this constraint. Since the highest score belonged to a school-level entry, the model's overly restrictive logic failed to retrieve the correct result.

\begin{table}[t!]
  \small
  \centering
  \begin{tabular}{p{0.15\linewidth} p{0.80\linewidth}}
  \toprule
  \textbf{Type} & \textbf{Content} \\
  \midrule
  \textbf{Question} & 
  Name schools in Riverside where the average math score is greater than 400, and what is the funding type? \\
  \midrule
  \textbf{Gold SQL} & 
  \texttt{SELECT T1.sname, T2.Funding FROM ...} \newline
  \texttt{WHERE T2.District LIKE 'Riverside\%'} \newline
  \textbf{\texttt{GROUP BY T1.sname, T2.Funding}} \newline
  \texttt{HAVING ... > 400} \\
  \midrule
  \textbf{Prediction \newline (Pandas)} & 
  \texttt{df\_merged = pd.merge(df\_sat, df\_frpm, ...)} \newline
  \texttt{\# Implicitly drops NaNs in grouping keys} \newline
  \textbf{\texttt{grouped = df\_merged.groupby(['sname', 'Funding'])}} \newline
  \texttt{result = grouped['AvgScrMath'].mean()} \\
  \bottomrule
  \end{tabular}
  \caption{A case study of semantic mismatch between SQL and Pandas. While the SQL \texttt{GROUP BY} retains \texttt{NULL} keys, the generated Pandas code uses the default \texttt{groupby} which drops \texttt{NaN} values, reducing recall.}
  \label{tab:mismatch_example}
\end{table}

\begin{table}[t!]
    \small
    \centering
    \begin{tabular}{p{0.15\linewidth} p{0.80\linewidth}}
    \toprule
    \textbf{Type} & \textbf{Content} \\
    \midrule
    \textbf{Question} & 
    Which cities have the top 5 lowest enrollment number for students in grades 1 through 12? \\
    \midrule
    \textbf{Gold SQL} & 
    \texttt{SELECT T2.City FROM ...} \newline
    \textbf{\texttt{ORDER BY SUM(T1.`Enrollment`) ASC LIMIT 5}} \newline
    \vspace{0.3em}
    \textit{Execution Result (Top 5):} \newline
    ['Coulterville', 'Pinecrest', 'Shaver Lake', \textbf{'Emigrant Gap', 'Hyampom'}] \\
    \midrule
    \textbf{Prediction \newline (Pandas)} & 
    \texttt{df\_merged = pd.merge(...)} \newline
    \textbf{\texttt{df\_sorted = df\_merged.sort\_values('Enrollment').head(5)}} \newline
    \vspace{0.3em}
    \textit{Execution Result (Top 5):} \newline
    ['Coulterville', 'Pinecrest', 'Shaver Lake', \textbf{'Hyampom', 'Woody'}] \newline
    \textit{(Note: 'Emigrant Gap' is excluded due to tie-breaking)} \\
    \bottomrule
    \end{tabular}
    \caption{Example of Nondeterministic Sort Stability. Both SQL and Pandas correctly sorted by enrollment, but the absence of a secondary sort key caused divergent results at the cutoff point ($N=5$) for tied values.}
    \label{tab:tie_breaking_mismatch}
\end{table}

\begin{table}[t]
    \small
    \centering
    \begin{tabular}{p{0.15\linewidth} p{0.80\linewidth}}
    \toprule
    \textbf{Type} & \textbf{Content} \\
    \midrule
    \textbf{Question} & 
    Which active \textbf{district} has the highest average score in Reading? \\
    \midrule
    \textbf{Gold SQL} & 
    \texttt{SELECT T1.District FROM schools ...} \newline
    \texttt{WHERE T1.StatusType = 'Active'} \newline
    \textbf{\texttt{ORDER BY T2.AvgScrRead DESC LIMIT 1}} \newline
    \vspace{0.3em}
    \textit{Result:} \textbf{Palo Alto Unified} (Score: 642.0, Type: S) \\
    \midrule
    \textbf{Prediction \newline (Pandas)} & 
    \texttt{\# Explicitly filtering for District-level records} \newline
    \textbf{\texttt{df\_scores = df\_sat[df\_sat['rtype'] == 'D']}} \newline
    \texttt{df\_merged = pd.merge(df\_schools, df\_scores, ...)} \newline
    \vspace{0.3em}
    \textit{Result:} \textbf{Santa Cruz County Office} (Score: 638.0, Type: D) \\
    \bottomrule
    \end{tabular}
    \caption{Example of Granularity Mismatch. The model inferred a schema constraint (\texttt{rtype='D'}) from the word "district", while the SQL ground truth operated on a global level, leading to divergent answers.}
    \label{tab:granularity_mismatch}
\end{table}

\section{Logic Completion Framework (LCF) Specifications}
\label{sec:appendix_lcp}

This section details the experimental configuration for the LCF roles and interaction pipeline.

\subsection{Role Configuration}
\textbf{Subject (Evaluated Model).} The Subject is responsible for both ambiguity detection (Phase 1) and code generation (Phase 3). Assigning the inquiry task to the Subject evaluates its intrinsic ability to identify missing information, avoiding dependence on an external model's detection capabilities.

\textbf{Oracle (Ground Truth Source).} The Oracle (Qwen3-max) functions as a domain expert. It utilizes the Ground Truth SQL/Code to address the Subject's inquiries. The Oracle provides natural language hints to guide reasoning without revealing the target code.

\subsection{Pipeline Specification}
The LCF process comprises three steps:
\begin{enumerate}
    \item \textbf{Logic Probing:} The Subject identifies ambiguities in the schema or task requirements and formulates a clarifying question ($Q_{ambiguity}$).
    \item \textbf{Constraint Generation:} The Oracle derives logic from the ground truth to produce natural language hints ($C_{logic}$) that answer the inquiry.
    \item \textbf{Execution:} The Subject generates the final program based on the original query and the provided constraints.
\end{enumerate}

\subsection{LCF Prompt Templates}
Table \ref{tab:lcp_phase1_prompt} and Table \ref{tab:lcp_phase2_prompt} present the system prompts for Phase 1 and Phase 2. Phase 3 uses the standard generation prompt (Appendix \ref{sec:appendix_prompts}) augmented with constraints from Phase 2.

\begin{table}[t!]
    \small
    \centering
    \begin{tabular}{p{0.95\linewidth}}
    \toprule
    \textbf{Phase 1 Prompt: Logic Probing (System Instruction)} \\
    \midrule
    -- As an expert Data Analyst, identify ambiguity in the following question regarding the data schema or business logic. \\
    -- Task: You are a logical analyst. Identify any ambiguity in the question regarding the database schema, business logic, or output structure requirements (e.g., exact columns to return, handling of NULLs). Return ONLY a clarifying question that addresses the ambiguity. Do not generate SQL or Code. \\
    \bottomrule
    \end{tabular}
    \caption{System prompt for the Subject model in Phase 1 (Logic Probing).}
    \label{tab:lcp_phase1_prompt}
\end{table}

\begin{table}[t!]
    \small
    \centering
    \begin{tabular}{p{0.95\linewidth}}
    \toprule
    \textbf{Phase 2 Prompt: Constraint Generation (Oracle Instruction)} \\
    \midrule
    -- Task: You are a Data Analyst Proxy. Based strictly on the Ground Truth provided below, answer the Model's Inquiry by explaining the business constraints or logic used in the Gold implementation. \\
    -- Ground Truth: \{GOLD\_QUERY\} \\
    -- Model's Inquiry: \{PHASE1\_INQUIRY\} \\
    Return your response in JSON format with "classification" and "answer". \\
    -- Constraints: \\
    1. Do NOT output code. \\
    2. Extract the business logic into implementation-agnostic natural language hints (e.g., "Filter by status 'Active'" instead of "WHERE status = 'Active'"). Do not leak implementation details or syntactic tokens. \\
    3. Answer ONLY what is explicitly asked in the Inquiry. \\
    4. Explicitly mention specific Column Names and Values required. \\
    \bottomrule
    \end{tabular}
    \caption{System prompt for the Oracle model in Phase 2 (Specification Generation).}
    \label{tab:lcp_phase2_prompt}
\end{table}

\subsection{LCF Dialogue Case Study}
We present a case study from the development set in Table \ref{tab:lcp_case_study} to illustrate the interaction within the LCF. This example shows how the protocol resolves schema ambiguity and computational constraints.

In Phase 1, the Subject identifies ambiguity regarding the definition of "continuation school" (specifically, the choice between \texttt{schools.StatusType} and \texttt{frpm.Educational Option Type}) and questions boundary conditions for arithmetic operations. In Phase 2, the Oracle provides implementation-agnostic constraints derived from the Ground Truth. It clarifies the correct table source and defines the logic for handling NULL values without revealing SQL syntax.

\begin{table}[t!]
    \small
    \centering
    \begin{tabular}{p{0.95\linewidth}}
    \toprule
    \textbf{Case Study: Eligible Free Rates Calculation} \\
    \midrule
    \textbf{Input Question} \\
    "Please list the lowest three eligible free rates for students aged 5-17 in continuation schools." \\
    \midrule
    \textbf{Phase 1 Output: Ambiguity Identification (Model Inquiry)} \\
    "What defines a 'continuation school' in the \texttt{StatusType} column of the \texttt{schools} table, and how should division by zero or NULL values be handled when calculating the eligible free rate?" \\
    \textit{Observation: The model correctly questions the calculation logic but holds an incorrect assumption about the schema location of the school type definition.} \\
    \midrule
    \textbf{Phase 2 Output: Constraint Generation (Oracle Response)} \\
    "A 'continuation school' is defined by the value 'Continuation School' in the \texttt{Educational Option Type} column of the \texttt{frpm} table, not by the \texttt{StatusType} column in the \texttt{schools} table. When calculating the eligible free rate for students aged 5–17, only records where \texttt{Free Meal Count (Ages 5-17)} is not NULL and \texttt{Enrollment (Ages 5-17)} is greater than zero are included, thereby avoiding division by zero and excluding undefined rates." \\
    \bottomrule
    \end{tabular}
    \caption{Example of the LCF workflow. The interaction moves from initial schema uncertainty to logic specification, facilitating code generation.}
    \label{tab:lcp_case_study}
\end{table}

\end{document}